\documentclass[11pt,a4paper]{article}
\usepackage[hyperref]{emnlp2018}

\usepackage{times}
\usepackage{latexsym}
\usepackage{multirow}
\usepackage{caption}
\usepackage{subcaption}
\usepackage{microtype}
\usepackage{arydshln}
\usepackage{algorithm2e}

\usepackage{url}

\PassOptionsToPackage{hyphens}{url}
\usepackage{hyperref}

\makeatletter
\g@addto@macro{\UrlBreaks}{\UrlOrds}
\makeatother

\usepackage{latexsym}
\usepackage{graphicx}
\usepackage{fixltx2e}
\usepackage{epstopdf}
\usepackage{amssymb}
\usepackage{amsmath}
\usepackage{xspace,relsize,gensymb}
\usepackage{graphics,boxedminipage}
\usepackage{adjustbox,paralist}
\usepackage{booktabs}
\usepackage{arydshln}
\usepackage{soul}
\usepackage{color}
\usepackage{adjustbox,booktabs,enumitem}
\usepackage[utf8]{inputenc}
\usepackage{pgfplots}

\newcommand{\class}[1]{\texttt{#1}\xspace}

\newcommand{\method}[2][]{\texttt{#2}$_{\text{#1}}$\xspace}

\newcommand{\cccm}{\method{CCCM}}

\newcommand{\ssa}{\method{SSA}}
\newcommand{\ssb}{\method{SSB}}
\newcommand{\nlp}{\method{NLP}}
\newcommand{\gpttwo}{\method{GPT2}}

\newcommand{\ex}[1]{\textit{#1}\xspace}

\newcommand{\secref}[2][]{Section#1~\ref{sec:#2}}

\newcommand{\tabref}[2][]{Table#1~\ref{tab:#2}}

\newcommand\email{\begingroup \urlstyle{tt}\smaller\Url}

\newenvironment{smallitemize}
{\begin{itemize}
  \setlength{\itemsep}{6pt}
  \setlength{\parskip}{0pt}
  \setlength{\parsep}{0pt}}
{\end{itemize}}

\aclfinalcopy 

\title{You are right. I am ALARMED -- But by Climate Change Counter Movement}

\author{Shraey Bhatia \qquad Jey Han Lau \qquad Timothy 
Baldwin \\[1ex]
 School of Computing and Information Systems\\The University 
   of
Melbourne \\[0.5ex]
  \email{shraeybhatia@gmail.com}, \email{jeyhan.lau@gmail.com}, 
 \email{tb@ldwin.net}}

\begin{document}

\maketitle

\begin{abstract}

The world is facing the challenge of climate crisis. Despite the consensus in scientific community about anthropogenic global warming, the web is flooded with articles spreading climate misinformation. These articles are carefully constructed by climate change counter movement (\cccm) organizations to influence the narrative around climate change. We revisit the literature on climate misinformation in social sciences and repackage it to introduce in the community of \nlp. Despite considerable work in detection of fake news, there is no misinformation dataset available that is specific to the domain.of climate change. We try to bridge this gap by scraping and releasing articles with known climate change misinformation. 
\end{abstract}

\section{Introduction}
\label{sec:intro}

Climate change is one of the biggest challenges threatening the world,
and we are at the defining moment. Rising sea levels, melting polar ice,
changing weather patterns, severe droughts, and extinction of species
are just some of the dreadful effects of this crisis. The
Intergovernmental Panel on Climate Change (IPCC) in its 5th assessment
report categorically concluded that humans are the main culprit and
there is a need to limit global warming to less than
2\degree C.\footnote{\url{https://www.ipcc.ch/report/ar5/wg1/}} More
recently, anthropogenic climate change has been at the heart of the
Australian bushfires \cite{nhess+:2020}, leading to the destruction of
17 million hectares of land and the death of a billion
animals.\footnote{\url{https://www.aph.gov.au/About_Parliament/Parliamentary_Departments/Parliamentary_Library/pubs/rp/rp1920/Quick_Guides/AustralianBushfires}}
During these times, we see articles with headlines such as \ex{Climate Change has
  caused more rain, helping fight Australian wildfires} spreading
misinformation to influence the narrative of climate
change.\footnote{\url{https://www.heartland.org/news-opinion/news/climate-change-has-caused-more-rain-helping-fight-australian-wildfires}}

The rise of this misleading information is part of a carefully crafted
strategy by climate change counter movement (\cccm) organizations
\cite{dunlap+:2013, boussalis+:2016, farrell+:2016b, mckie+:2018}. These
organizations use a formula consisting of a narrative structured around
the principle ingredients of disinformation, misinformation, propaganda
and hoax, sprinkled with the stylistic elements of sensationalism,
melodrama, clickbait and satire, as can be seen in examples in
\tabref{examples}. Their approach broadly mirrors that seen in
fake news in the political arena \cite{rashkin+:2017}, but is specifically
tailored to the domain of climate change. This motivates the development
of applications that can inform users via an automatic detection or 
alert system, similar to what we have seen for fake news
\cite{rashkin+:2017, perez+:2017, jiang+:2018}.

\begin{table}[t]
\begin{center}
\begin{tabular}{p{\dimexpr \linewidth-2\tabcolsep}}
\toprule
German evolutionary biologist and physiologist Prof.\ Dr.\ Ulrich 
Kutschera told in an interview that CO$_{2}$ is a blessing for mankind 
and that the claimed 97\% consensus among scientists is a myth. ...
he rejected extremes, among them the climate alarmists who predict 
a fictitious, imminent earth heat death and thus practice a kind of 
religious cult. \\
\midrule
New Zealand schools to terrify children about the climate crisis. Who 
cares about education if you believe the world is ending? What will it 
take for sanity to return? Global cooling? Another Ice Age even? The 
climate lunatics ...  encourage them to wag school to protest for more 
action.\\
\bottomrule
\end{tabular}
\end{center}
\caption{Articles created by \cccm organizations.}
\label{tab:examples}
\end{table}

%
%
%

The lack of annotated fake news data spurred the creation of
misinformation datasets. The first public dataset for fake news
detection \cite{vlachos+:2014} and claim/stance verification
\cite{ferreira+:2016} are moderately small with 221 and 300 instances,
respectively. More recently, larger datasets have been developed, such as LIAR
\cite{wang+:2017}, collected from PolitiFact and labelled with 6 levels
of veracity, and FEVER \cite{thorne+:2018}, a dataset generated from
Wikipedia with \class{supported}, \class{refuted} and \class{not
  enough info} labels.  Extending the task to full articles,
FakeNewsNet is a valuable resource \cite{shu+:2017, shu+:2018}.  But to
the best of our knowledge there is no misinformation dataset that is
specific to the domain of climate change.  We attempt to fill this gap
by releasing a large set of documents with known climate change
misinformation.



\section{Related Work}
\label{sec:related-work}

The way the public perceives and reacts to the constant supply of 
information around climate change is a function of how the facts and 
narrative are presented to them \cite{flottum+:2014, flottum+:2016}.
\newcite{flottum+:2017} emphasizes that language and communication 
around climate change are significant, as climate is not just the physical 
science but has political, social and ethical aspects, and involves 
various stakeholders, interests and voices.  A range of corpus 
linguistic methods have been used to study the topical and 
stylistic aspects of language around climate change.  
\newcite{tvinnereim+:2015} proposed the use of structured topic 
modelling \cite{roberts+:2014} to derive insights about the public 
opinion from 2115 open-ended survey responses.  \newcite{salway+:2014} 
leveraged unsupervised grammar induction and pattern extraction methods to 
find common phrases in climate change communication.  
\newcite{atanasova+:2017} analysed frequently-used metaphors manually in 
editorials and op-eds, and concluded that the communication in 
the Guardian (U.K.) was predominantly  war based 
(e.g.\ \ex{threat of climate change}), Seuddeutsche
(Germany) based on illness (e.g.\  \ex{earth has fever}), and 
the NYTimes (U.S.A) based on the idea of a journey 
(e.g.\ \ex{many small steps in the right direction}). 

In linguistics, style broadly refers to the properties of a sentence
beyond its content or meaning \cite{pennebaker+:1999}, and stylistic
variation plays an important role in the identification of
misinformation.  \newcite{biyani+:2016} studied stylistic aspects of
clickbait and formalised it into 8 different categories ranging from
exaggeration to teasing, and proposed a clickbait classifier based on
novel informality features.  Similarly, \newcite{kumar+:2016}
examined the unique linguistic characteristics of hoax documents in
Wikipedia and built a classifier using a range of hand-engineered
features.  \newcite{rashkin+:2017} proposed using stylistic lexicons
(e.g.\ Linguistic Inquiry and Word Count (LIWC)), subjective words, and
intensifying lexicons for fact checking, and demonstrated that words used
to exaggerate like superlatives, subjectives, and modal adverbs are
prominent in fake news, whereas trusted sources are dominated by
assertive words. \newcite{wang+:2017} experimented with detecting fake
news using meta data features with convolutional neural networks adapted
for text \cite{kim+:2014}.

Although articles with misinformation are predominantly human-written,
the recent emergence of large pre-trained language models means they can
now be automatically generated.  \newcite{radford+:2019} introduced a
large auto-regressive model (\gpttwo) with the ability to generate
high-quality synthetic text. One limitation of \gpttwo is its inability
to perform controlled generation for a specific domain, and
\newcite{keskar+:2019} proposed a model to tackle this.  Building on
this further, \newcite{dathathri+:2019} introduced a plug and play
language model, where the language model is a pretrained model similar to \gpttwo but with controllable components that can be fine-tuned through attribute classifiers.

\section{Climate Change Counter Movement Organisations}
\label{sec:cccm-intro}

Despite the findings of the IPCC's 5th Assesment Report and more than 97
percent consensus in the scientific community to support anthropogenic
global warming \cite{cook+:2013}, coordinated efforts to tackle the
climate crisis are lacking. This can be attributed to the rise in
opposing voices including the fossil fuel lobby, conservative
thinktanks, big corporations, and digital/print media questioning the
science and research around climate change. These organisations are
collectively referred to as climate change counter movement (``\cccm'')
organizations \cite{oreskes+:2010, dunlap+:2013, farrell+:2016b,
  boussalis+:2016, mckie+:2018}.  \newcite{mckie+:2018} argues that the
motivation behind these organizations is to maintain the status quo of
the hegemony of fossil fuel-based neo-liberal global capitalism. These
organizations are found around the globe and can masquerade as
philanthropic organizations to fund climate misinformation
\cite{farrell+:2019}, hide behind libertarian ideas \cite{mckie+:2018}
to question the scientists, and augment scepticism to promote pseudo
science or `alternative facts'.  Some of these organizations have catchy
names such as \url{carbonsense.com} or \url{friendsofscience.org}, and
organize their own scientific conferences.

\newcite{oreskes+:2010} concluded in their analysis that the strategies
employed by \cccm to construct the narrative to spread misinformation 
resembles the ones historically used by the tobacco lobby. For 
instance, targeting researchers and questioning the methodology of their 
research, and blaming scientific standards are consistent strategies used 
by both \cccm and tobacco lobby groups \cite{oreskes+:2010,mckie+:2018}.  
\newcite{dunlap+:2015,farrell+:2016b, boussalis+:2016} categorized their 
misinformation arguments into 2 frames: science and policy.  Science 
frame arguments question the scientific facts and deliberately plant a 
lie to sway the public towards pseudo science, whereas policy frame 
arguments target issues of cost and economy (e.g.\ carbon tax) or pass 
the blame for action to other nations.  We present several examples of 
arguments in the science and policy frames in \tabref{sp-examples} .

\begin{table}[t]
\centering
\begin{adjustbox}{max width=\linewidth}
\begin{tabular}{ll}
\toprule
\textbf{Argument} & \textbf{Frame} \\
\midrule
CO2 is plant food and is good for the planet             & Science \\ 
Climate change is natural and has always been changing   & Science \\
We are entering another ice age                          & Science \\ 
Adapting to global warming is cheaper than preventing it & Policy  \\ 
Renewable energy is way too expensive                    & Policy  \\ 
\bottomrule
\end{tabular}
\end{adjustbox}
\caption{Examples of counter climate arguments and their frames.}
\label{tab:sp-examples}
\end{table}

We believe the narrative of \cccm articles have two aspects: topical and 
stylistic. The topical aspects describe common issues discussed in \cccm 
articles (e.g.\ carbon tax, fossil fuel, and renewable energy). The 
stylistic aspects capture how the narrative is presented --- e.g.\ the 
use of exaggeration and sensationalism, as evident in the examples in 
\tabref{examples} --- and bear similar characteristics to fake news.


\section{Dataset}
\label{sec:datasets}

To construct our dataset, we scrape articles with known climate change 
misinformation from 15 different \cccm organizations.  These 
organizations are selected from three sources: (1) \newcite{mckie+:2018},
(2)  
\textit{desmogblog.com},\footnote{\url{https://www.desmogblog.com/global-warming-denier-database}} 
a website that maintains a database of individuals and organizations 
that have been identified to perpetuate climate disinformation; and (3) an organizations cited on the website selected from above 2 sources. A number of considerations were made when developing the dataset:

\begin{itemize}
\item We only scrape articles from organizations active in
  English-speaking countries: the United States, Canada, United Kingdom,
  Australia, and New Zealand.

\item A considerable number of organizations are either dormant or have
  a very low level of activity. To make sure our dataset is up to date,
  we only scrape articles from organizations with a reasonable level of
  activity, e.g.\ they publish at least 1 article every month, and their
  latest publication is in 2020.

\item We set a minimum and maximum threshold of 10 and 400 articles
  respectively for each organization. We set a maximum threshold so as
  to avoid bias towards one organization. Note, however, that there is a
  considerable variance in the article length for different
  organizations.  For instance, one organization with only 10 articles
  has an average length of 342.1 words, while another organization with
  400 articles has an average length of 85.8 words.

\item As explained in \secref{cccm-intro}, counter climate arguments can
  be broadly categorised into the science and policy frames. As
  organizations generally prefer one type of frame in their
  narrative, we manually identify frames associated with organizations, and
  select a set of organizations that produces a balanced representation
  of both frames in the dataset.
\end{itemize}

We split the documents into training and test partitions at the 
organization level, where the training set comprises 12 organizations 
and the test set 3 organizations. We split at the organization level 
because  it allows us to test whether detection models are able to 
generalize their predictions to articles published by unseen/new \cccm 
organisations.
We present some statistics for the training documents in 
\tabref{train-stats}, and the list of organizations is provided in the 
supplementary file.

\begin{table}[t]
\centering
\begin{adjustbox}{max width=1.0\textwidth}
\begin{tabular}{rl}
\toprule
\# Organizations & 12    \\
\# Articles      & 1168  \\
Mean Length      & 559.3 \\
Median Length    & 332.0   \\
Std Length  & 640.2  \\
\bottomrule
\end{tabular}
\end{adjustbox}
\caption{Statistics of \cccm training data}
\label{tab:train-stats}
\end{table}

\subsection{Test data}
\label{sec:test-set}

We extend our test set to include documents that do not have climate
change misinformation, to create a standard evaluation dataset for
climate change misinformation detection.  We collect documents from
reputable sources, some of which are not climate-related, and some are
satirical in nature.  Sources of the full test documents are as
follows:
\begin{smallitemize}

\item \textbf{Guardian}: A trusted source for independent journalism.
  We scrape articles under the category of climate change from both its
  U.K.\ and Australian editions. These articles test whether a detection
  system can correctly identify these articles as not having climate
  change misinformation.

\item \textbf{BBC}: Similar to Guardian, we scrape articles from their
  website under the category of climate change.

\item \textbf{Newsroom}: This is a dataset released by
  \newcite{grusky+:2018} and consists of articles and summaries compiled
  from 38 different publications.\footnote{\url{https://summari.es/}} We
  take a random sample of articles which are not climate related.  These
  articles test whether a detection system is able to identify
  non-climate-related articles as not having climate change
  misinformation.

\item \textbf{Beetota Advocate}: This is an Australian satirical website
  which publishes articles on current affairs happening locally and
  internationally;\footnote{\url{https://www.betootaadvocate.com/}} we
  scrape articles related to climate change. Although there is a tone of
  sensationalism in the writing, the articles are created with the
  intent of humour.  An example of a Beetota Advocate article is given
  in \tabref{beet-examples}.  These articles test whether detection
  models are able to distinguish them from \cccm articles, as both have
  similar stylistic characteristics.

\item \textbf{Sceptical Science Arguments (\ssa) and Sceptical Science
    Blogs (\ssb)}: This resource focuses on explaining what science says
  about climate
  change.\footnote{\url{https://www.skepticalscience.com/}} It publishes
  general climate blogs and counters common climate myths by putting
  forth arguments backed by peer-reviewed research.

\item \textbf{\cccm}: These are articles from the 3 \cccm organizations,
  as detailed in \secref{datasets}. These documents are the only
  documents with climate change misinformation in the test data.
\end{smallitemize}

\tabref{test-set} contains statistics of the test set.

\begin{table}
\begin{center}
\begin{tabular}{p{\dimexpr \linewidth-2\tabcolsep}}
\toprule
PM Meets With Cricket Side To Discuss The 1.7m Hectares Of NSW Forests 
Destroyed By Bushfires. Not even six months after being officially 
elected as the Australian Prime Minister with absolutely no policies, 
let alone any acknowledgement of his government’s denialism-led inaction 
on climate change, Scott Morrison has today had the opportunity to meet 
some more sportsmen! While the drought-stricken communities of rural 
Australian continue to burn at the hands of record-breaking and 
out-of-control bushfires, ScoMo has today met with the Australian 
cricket side for his ideal media appearance.. ...... The cricketers 
appeared distressed while also having to pose for goofy photos with the 
Prime Minister, ...... planet’s temperature that will result in the 
certain deaths of the billions of people that haven’t been given 
permission to join Gina Rinehart and her Liberal Party employees in the 
spaceship. \\
\bottomrule
\end{tabular}
\end{center}
\caption{A Beetota Advocate article on climate change.}
\label{tab:beet-examples}
\end{table}

\begin{table}[t]
\centering
\begin{adjustbox}{max width=\linewidth}
\begin{tabular}{cc}
\toprule
\textbf{Source} & \textbf{\#Doc} \\
\midrule
Guardian & 80 \\
BBC & 60 \\
Beetota & 70 \\
Newsroom & 100 \\
\ssa + \ssb & 50 \\
\cccm & 150 \\
\bottomrule
\end{tabular}
\end{adjustbox}
\caption{Test set document statistics.}
\label{tab:test-set}
\end{table}

\section{Conclusion}
\label{sec:conclusion}

We introduced climate misinformation to the domain of fake news and in the community of \nlp. We explored the literature around emergence of \cccm organizations, the strategies and linguistic elements used by these organizations to construct a narrative of climate misinformation. To help in countering its spread, we scrape articles with known sources of misinformation and release it to the community.



\bibliographystyle{acl_natbib}
\bibliography{acl2018,papers}

\end{document}


\begin{table*}[t]
\centering
\begin{adjustbox}{max width=1.0\textwidth}
\begin{tabular}{|l|l|}
\hline
Organization Name                     & Website                                \\ \hline
Australian Environmental Foundation   & https://www.australianenvironment.org/ \\ \hline
Australian Climate Madness            & https://australianclimatemadness.com/  \\ \hline
The Carbon Sense Coalition            & https://carbon-sense.com/              \\ \hline
CO2 Coalition                         & https://co2coalition.org/              \\ \hline
Fraser Institute                      & https://www.fraserinstitute.org/       \\ \hline
Friends of Science                    & https://friendsofscience.org/          \\ \hline
The Global Warming Policy Foundation  & https://www.thegwpf.org/               \\ \hline
Heartland Institute                   & https://www.heartland.org/             \\ \hline
The Institute of Public Affairs       & https://ipa.org.au/                    \\ \hline
Manhattan Institute                   & www.manhattan-institute.org/‎          \\ \hline
New Zealand Climate Science Coalition & https://www.climatescience.org.nz/     \\ \hline
Watts up with that                    & https://wattsupwiththat.com/           \\ \hline
Cato Institute                        & https://www.cato.org/                  \\ \hline
The BFD                               & https://thebfd.co.nz/                  \\ \hline
Heritage Foundation                   & https://www.heritage.org/              \\ \hline
\end{tabular}
\end{adjustbox}
\caption{Table of \cccm Organisations}
\label{tab:my-table}
\end{table*}